\documentclass[letterpaper, 10 pt, conference]{ieeeconf}  %

\IEEEoverridecommandlockouts                              %

\usepackage{graphics} %
\usepackage{epsfig} %
\usepackage{amssymb}  %
\usepackage{bm}
\usepackage{algpseudocode}
\usepackage{algorithm}
\usepackage{xcolor}
\usepackage{amsmath}
\usepackage{lineno}
\makeatletter
\let\NAT@parse\undefined
\makeatother
\usepackage[colorlinks=true, citecolor=green, linkcolor=red]{hyperref} 
\usepackage{multirow}
\usepackage{booktabs}
\usepackage{cite}
\usepackage{caption}
\usepackage{subcaption}

\usepackage{bbm}

\title{\LARGE \bf
Parameter-efficient Prompt Learning for 3D Point Cloud Understanding
}

\author{Hongyu Sun, Yongcai Wang, Wang Chen, Haoran Deng and Deying Li%
  \thanks{All authors are with the Department of Computer Science, School of Information, Renmin University of China, Beijing 100872, China. Corresponding author: Yongcai Wang (ycw@ruc.edu.cn). }
  \thanks{This work was supported in part by the National Natural Science Foundation of China under Grants 
  No. 61972404 and No. 12071478, and Public Computing Cloud, Renmin University of China, and the Blockchain
  Lab, School of Information, Renmin University of China.}
}

\begin{document}

\maketitle
\thispagestyle{empty}
\pagestyle{empty}

\begin{abstract}

This paper presents a parameter-efficient prompt tuning method, named PPT, 
to adapt a large multi-modal model for 3D point cloud understanding. 
Existing strategies are quite expensive in computation and storage, and depend on time-consuming prompt engineering. 
We address the problems from three aspects. Firstly, a PromptLearner module is devised to replace hand-crafted prompts with 
learnable contexts to automate the prompt tuning process. Then, we lock the pre-trained backbone instead of adopting the full fine-tuning 
paradigm to substantially improve the parameter efficiency. Finally, a lightweight PointAdapter module is arranged near target tasks to enhance 
prompt tuning for 3D point cloud understanding. 
Comprehensive experiments are conducted to demonstrate the superior parameter and data efficiency of the proposed method. 
Meanwhile, we obtain new records on 4 public datasets and multiple 3D tasks, i.e., point cloud recognition, few-shot learning, 
and part segmentation. 
The implementation is available at \url{https://github.com/auniquesun/PPT}. 

\end{abstract}

\section{Introduction}
\label{sec:intro}
Point cloud understanding plays a crucial role in real-world perception since the point cloud 
data is one of the most direct forms generated by 3D measuring equipment. 
Previously, PointNet~\cite{qi17pointnet} and PointNet++~\cite{qi17pointnet2} sparked a wave 
of directly operating irregular point clouds via deep learning-based architectures. 
After rapid progress for years~\cite{li18pointcnn,wang19dgcnn,thomas19kpconv,wu19pointconv,
yan20pointasnl,zhao20exploring,ran22repsurf,liu19rscnn,guo2020pct,zhao21pt,xiang21curvenet,ma22pointmlp,qian22pointnext},
the performances of point-based methods gradually approach a ceiling, 
partly due to the lack of texture and visual semantics in point cloud data, 
which are vital for many applications,
such as 3D object recognition, segmentation and detection. 

Inspired by the great success of large models in language and image understanding~\cite{devlin19bert,brown20gpt3,raffel20t5,radford21clip,jia21scaling,dosovitskiy21vit,liu21swin,li22blip,dehghani23scaling,li23blip2}, 
researchers attempt to transfer the rich textual and visual knowledge encoded in the foundation models to 
boost point cloud understanding~\cite{zhang22pointclip,zhu23pointclip2,dong23act,qi23recon,zeng23clip2}. 
Recently, ULIP~\cite{xue23ulip} learns a unified representation for language, image, and point cloud by contrastive pre-training 
on a large-scale triplet dataset derived from ShapeNet~\cite{shapenet2015}. 
After pre-training, the point cloud encoder 
has absorbed textual and visual information, then it is deployed by full fine-tuning on downstream tasks, 
such as 3D object classification and retrieval. 
Extensive experiments show ULIP achieves consistent gains over different point cloud architectures
(i.e., PointNet++~\cite{qi17pointnet2}, PointMLP~\cite{ma22pointmlp}, PointBERT~\cite{yu22pointbert}, PointNeXt~\cite{qian22pointnext}). Therefore, ULIP can be regarded as a large multi-modal model for 3D understanding. 

\begin{figure}[t]
  \begin{center}
    \includegraphics[width=\linewidth]{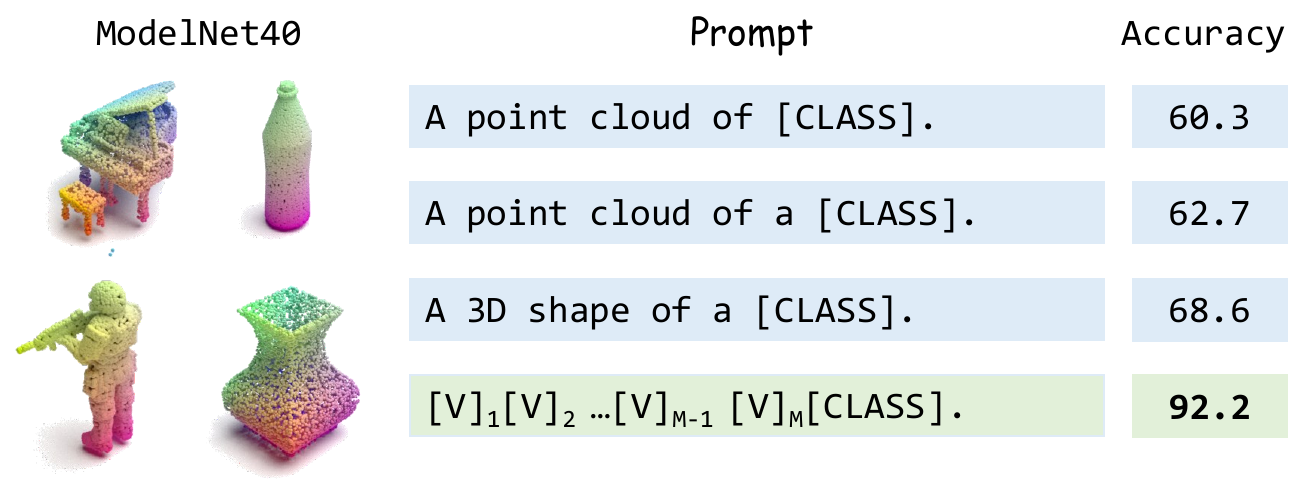}
    \includegraphics[width=\linewidth]{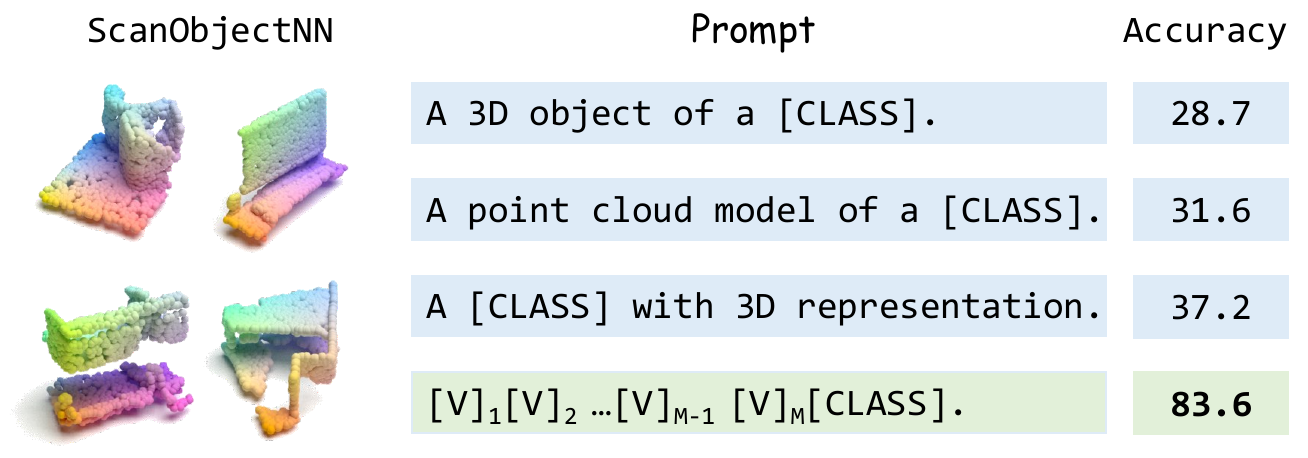}
  \end{center}
  \caption{Manual Prompts vs. Learnable Contexts. The former needs to find proper prompts manually. The latter 
  learns context vectors adaptively. The accuracy scores are obtained by running ULIP~\cite{xue23ulip} (PointBERT as 3D encoder).}
  \label{fig:manual_vs_soft_prompts}
\end{figure}

However, fully fine-tuning the pre-trained ULIP on downstream tasks is quite expensive and time-consuming  
since we need to update and store a separate copy of all parameters in the point encoder for each application. 
It is expected that there is a parameter-efficient way to leverage the power of ULIP.
Besides, we observe that prompt engineering in ULIP causes a fluctuation problem: a slight change to the hand-crafted prompts 
could have a big impact on performance, shown in Fig.~\ref{fig:manual_vs_soft_prompts}. 
For example, on the ModelNet40 dataset, when replacing ``a point cloud of a" with ``a 3D shape of a", the recognition accuracy increases by 5.9\%. 
Instead, when dropping the word ``a" from ``a point cloud of a", the accuracy decreases by 2.4\%.
In short, identifying a proper prompt manually is a non-trivial task. Sometimes, it requires domain expertise, 
but the result may be far from an optimal solution. 

To overcome the above problems, in this paper, we present an efficient and effective prompt tuning solution for 3D point cloud understanding, 
which not only improves parameter and data efficiency greatly but also exhibits better performances compared to  
strong baselines. 
Our solution is built on the recently released ULIP framework since it attains state of the art in several 3D tasks. 
Firstly, a PromptLearner module is devised to replace the hand-crafted prompts with learnable contexts. 
This design allows for finding proper prompts in vector space adaptively thus automating the prompt tuning process. 
Secondly, in contrast to the full fine-tuning paradigm, we lock the 3D encoder and prevent the parameters from updating. 
The strategy considerably improves parameter efficiency and saves computation and storage. 
Thirdly, we introduce a lightweight PointAdapter to further strengthen the performances of prompt tuning on downstream point cloud understanding tasks. 

To verify the effectiveness of the proposed method, we conduct various 3D perception tasks, 
i.e., standard point cloud recognition, few-shot classification, and 3D object part segmentation on four public datasets. 
The datasets vary from synthetic 3D objects~\cite{wu15modelnet,shapenet2015} to scanned scenarios in real world~\cite{uy19sonn}. 
The results reveal the excellent efficiency and effectiveness of PPT. 
In particular,
for point cloud recognition, our method reaches 94.1\% overall accuracy on the whole test set of ModelNet40~\cite{wu15modelnet} 
using only 1.8M learnable parameters and 30\% training data. 
Note that ULIP achieves the same performance with 
39.1M parameters and 100\% training samples. 
On the hardest split of ScanObjectNN~\cite{uy19sonn}, the proposed approach gets 89.1\% recognition accuracy, a 2.7\% absolute
improvement over ULIP, but requires only 50\% training data. 
For few-shot classification, our method demonstrates consistent advantages on two widely used datasets, especially
leading the runner up PointCLIP V2~\cite{zhu23pointclip2} by 19\% in the 16-shot setting of ScanObjectNN. 
For 3D part segmentation, PPT obtains 86.4 mean class IoU, which is a new record on ShapeNetPart~\cite{shapenetpart} 
while reducing the learnable parameters by 60\% compared to 
prior best method. 

In summary, the contributions of this paper include 
\begin{itemize}
  \item We identify two critical problems in ULIP: (1) performance fluctuation caused by prompt engineering. 
  (2) expensive storage and poor parameter efficiency caused by fully fine-tuning the pre-trained 3D encoder. 
  \item We devise PromptLearner and PointAdapter to liberate prompt engineering, promote parameter and data efficiency, 
  and enhance the effectiveness of point cloud understanding.
  \item The proposed method shows
  stunning performances across different tasks and datasets for 3D point cloud understanding, 
  supported by systematic experiments and ablation studies. 
\end{itemize}

\section{Related Work}
\label{sec:related_work}
Our work is related to developing a cheaper and easier-to-use prompt tuning strategy to
adapt a powerful multi-modal model to enrich point cloud understanding.

\noindent\textbf{Large Multi-Modal Models for 3D Tasks.} 
In recent years, large multi-modal models have shown incredible capabilities in text and
image understanding~\cite{radford21clip,jia21scaling,li22flip,li22blip,li23blip2}. 
Most of these models emphasize the interaction between text and image but lack 3D knowledge. 
A natural idea is to transfer the knowledge of powerful large models to promote 3D tasks. 
PointCLIP~\cite{zhang22pointclip} successfully achieved open-vocabulary 3D object recognition via projecting point clouds 
into images then exploiting the power of CLIP. 
PointCLIP V2~\cite{zhu23pointclip2} improved the predecessor by generating more realistic
projections and detailed descriptions for 3D objects. 
ACT~\cite{dong23act} explored the 3D representation learning assisted with pre-trained image/language models 
and demonstrated the benefits. 
I2P-MAE~\cite{zhang23i2pmae} proposed image-to-point mask auto-encoders to utilize pre-trained 2D models 
for 3D learning. 

Note that the above methods leverage powerful multi-modal models by converting point clouds into images or building an 
intermediate representation for 3D data. They don't touch the limitation of small scale and poor diversity of 
existing 3D datasets, which may deserve more attention. 
Recently, another branch of work has taken important steps toward this direction. 
The emergence of 3D datasets like Objavarse~\cite{Objaverse}, OmniObject3D~\cite{OmniObject3D} and ScanNeRF~\cite{ScanNeRF} 
greatly alleviated this limitation. Based on that, ULIP series~\cite{xue23ulip,xue23ulip2} quickly created 
large-scale text, image, point cloud triplets to learn a unified representation for the three modalities, then 
transferred the model to specific 3D tasks. 
CLIP$^2$~\cite{zeng23clip2} constructed million-scale triplets for contrastive pre-training and enhancing the 
generalization of learned 3D representations. 

Our goal is not to develop another large multi-modal model for point cloud understanding. Instead, we aim to substantially optimize 
the parameter and data efficiency of existing large models since current ways are expensive in storage and computation. 
Thus, this work is orthogonal to related work. 

\noindent\textbf{Prompt Learning for Large Models.}
The basic idea of prompt learning is to provide the model with task-related descriptions to elicit the knowledge learned in the pre-training stage 
rather than updating parameters in the backbone. 
This topic was originally investigated in NLP~\cite{shin20autoprompt,jiang20how,li21prefix,zhong21factual,lester21power,
gao21making,liu2021gpt,ben22bitfit} to adapt pre-trained large language models~\cite{brown20gpt3,devlin19bert} to downstream tasks. 
Since it only needs to optimize the text descriptions in the inputs while keeping the backbone untouched, 
and the results are promising in many applications, the strategy is quickly introduced in tuning
vision~\cite{jia22vpt,huang23damvp,zhang23cafo} and vision-language models~\cite{zhou22coop,zhou22cocoop,khattak23maple,khattak23promptsrc}.

However, a point cloud is an irregular structure consisting of sparse and unordered points, which essentially differs from text and image data. 
It is still unclear whether the parameter-efficient tuning strategy is effective for point cloud understanding. 
In this paper, we explore this problem by designing PromptLearner and PointAdapter modules based on recently released 
multi-modal framework ULIP~\cite{xue23ulip,xue23ulip2}, aiming at making ULIP-based model cheaper and easier for 3D
point cloud understanding. 

\begin{figure*}[t]
  \begin{center}
    \includegraphics[width=0.98\linewidth]{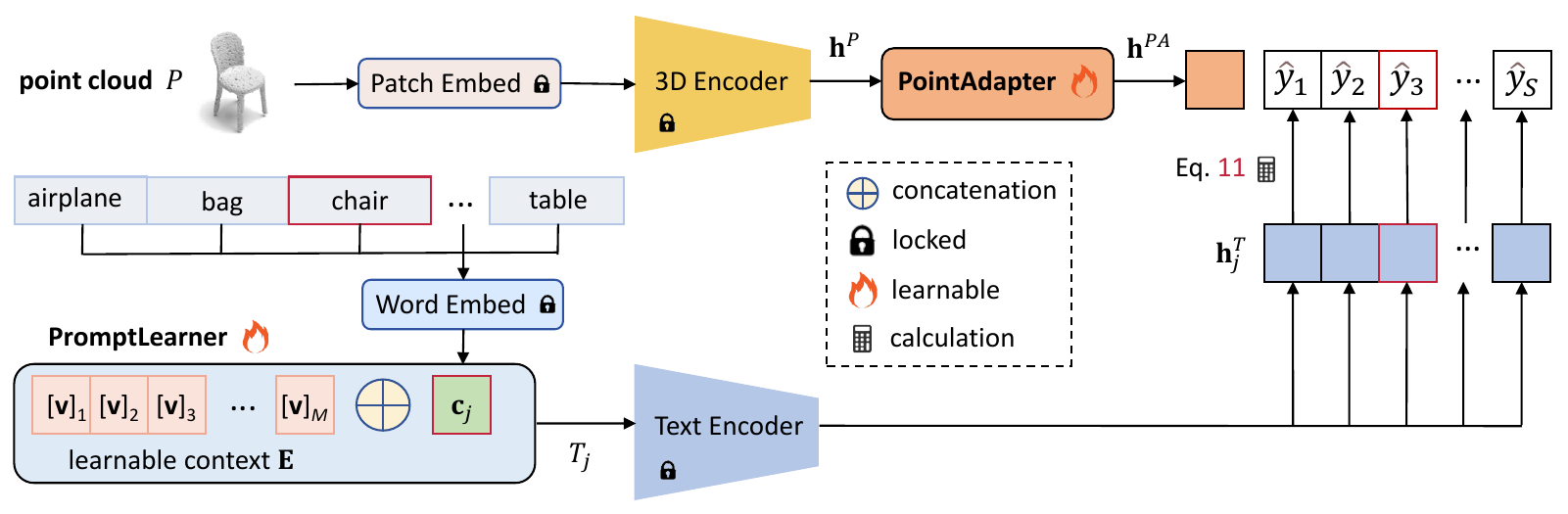}
  \end{center}
  \caption{The overall architecture of the proposed method. The class name embedding $\textbf{c}_j$ can be inserted in any position of the learnable vectors. Here we insert it in the end for illustration.}
  \label{fig:architecture}
\end{figure*}

\section{Methodology}
\label{sec:methodology}

In Section \ref{subsec:revisit_ulip}, we firstly recap the ULIP framework that forms the basis of the proposed PPT. 
Then in Section \ref{subsec:learn_to_prompt_ulip}, the details of the parameter-efficient prompt tuning method are elaborated. 

\subsection{Revisiting ULIP}
\label{subsec:revisit_ulip}

One highlight of ULIP is the construction of a large-scale text, image, and point cloud triplet dataset. 
For a triplet $U_i = (I_i, T_i, P_i)$, we denote the image as $I_i$, text as $T_i$ and point cloud as $P_i$. 
The corresponding encoders for the three modalities are $f_I(\cdot)$, $f_T(\cdot)$ and $f_P(\cdot)$, respectively.
Hence, the extracted features for $U_i$ can be represented as 
\begin{equation}
  \textbf{h}_i^{I} = f_I(I_i), \quad
  \textbf{h}_i^{T} = f_T(T_i), \quad
  \textbf{h}_i^{P} = f_P(P_i)
\end{equation}

Then ULIP learns a unified representation for the three modalities through unsupervised pre-training. 
The objective to be optimized is a contrastive loss, as in Eq.~\ref{eq:two_modality_contrast_loss}. %
\begin{equation}
  \begin{split}
  \mathcal{L}_{(M_1, M_2)} = \sum_{(i,j)} &-\frac{1}{2} \log\frac{\exp({s(\textbf{h}_i^{M_1},\textbf{h}_j^{M_2})})}{\sum_k \exp({s(\textbf{h}_i^{M_1},\textbf{h}_k^{M_2})})} \\
  &-\frac{1}{2} \log\frac{\exp({s(\textbf{h}_i^{M_1},\textbf{h}_j^{M_2})})}{\sum_k \exp({s(\textbf{h}_k^{M_1},\textbf{h}_j^{M_2})})}
  \end{split}
  \label{eq:two_modality_contrast_loss}
\end{equation}
where $M_1$ and $M_2$ are two different modalities, $(i,j)$ indexes a positive pair in a batch of training data, and
$s(\cdot,\cdot)$ computes the cosine similarity of the inputs. Therefore, the total loss of three modalities
can be computed by Eq.~\ref{eq:total_contrast_loss}.
\begin{eqnarray}
  \mathcal{L}_{total} = \alpha\mathcal{L}_{(I, T)} + \beta\mathcal{L}_{(I, P)} + \theta\mathcal{L}_{(P, T)}
  \label{eq:total_contrast_loss}
\end{eqnarray}

The weights of $f_I$ and $f_T$ are initialized with the vision-language model SLIP~\cite{mu21slip} then frozen. 
During pre-training, ULIP only updates the 3D encoder $f_P$.
After that, $f_P$ that absorbs 3D, textual, and visual knowledge is transferred to downstream 3D tasks by full fine-tuning. 

\subsection{Parameter-efficient Prompt Tuning for ULIP}
\label{subsec:learn_to_prompt_ulip}

Although ULIP refreshed records on multiple downstream tasks, including zero-shot point cloud recognition and standard 3D classification, 
the prompt engineering and full fine-tuning hinder it from being exploited easily and efficiently. We address the shortcomings
by designing the following modules: PromptLearner and PointAdapter. The overall pipeline of our method is presented in 
Fig.~\ref{fig:architecture}. 

\subsubsection{From Prompt Engineering to PromptLearner} 
ULIP generates text descriptions for a 3D point cloud through hand-crafted templates (i.e., ``a point cloud model of \verb|[CLASS]|'') then feeds them into the text encoder. 

In contrast, we develop PromptLearner to replace manual prompts with learnable contexts. 
The basic idea is to provide the text encoder with rich contexts adaptively rather than using fixed descriptions, 
to facilitate the model activating related knowledge. 
Specifically, the learnable contexts $\textbf{E}$ consist of $M$ continuous vectors, formulated in Eq.~\ref{eq:soft_prompts}, 
where $\textbf{v}_i \in \mathbb{R}^D$, $i = 1\dots M$. 
Note that they have the same form as the word embeddings of manual prompts. 
\begin{equation}
  \textbf{E} = [\textbf{v}]_1[\textbf{v}]_2\cdots[\textbf{v}]_M
  \label{eq:soft_prompts}
\end{equation}

For a downstream dataset of $S$ object categories for recognition, we concatenate the learnable contexts
$\textbf{E}$ with the word embedding of $j$th category name $\textbf{c}_j \in \mathbb{R}^{D}$, to 
generate the encoding $T_j \in \mathbb{R}^{(M+1)\times D}$. Then $T_j$ is fed into the text encoder to produce
the text feature $\textbf{h}_{j}^{T} \in \mathbb{R}^{D}$.
The procedure is formulated by Eq.~\ref{eq:input_of_text_encoder} and Eq.~\ref{eq:output_of_text_encoder}.
\begin{equation}
  T_j = [\textbf{E}, \textbf{c}_j], \quad j=1\dots S
  \label{eq:input_of_text_encoder}
\end{equation}
\begin{equation}\abovedisplayskip=0pt
  \textbf{h}_{j}^{T} = f_T(T_j), \quad j=1\dots S
  \label{eq:output_of_text_encoder}
\end{equation}

Hence, the text features are derived from learnable context vectors instead of fixed word embeddings. 
The optimization objective will be explained later. 

\subsubsection{From Full Fine-tuning to PointAdapter}
We discard expensive full fine-tuning and switch to learning a lightweight PointAdapter, denoted as $f_{PA}(\cdot)$. 
The module is arranged after the 3D encoder to enhance the performances of downstream 3D tasks. 

In the 3D branch, we lock the point patch embedding module and 3D encoder while ensuring the parameters in PointAdapter 
are updatable. 
A point cloud $P \in \mathbb{R}^{N\times 3}$ is processed with the embedding and 3D encoder to obtain the 
representation $\textbf{h}^{P} \in \mathbb{R}^{D}$, where $N$ is the number of points. 
Then $\textbf{h}^{P}$ is further processed by our PointAdapter to produce new representation $\textbf{h}^{PA}$ 
to adapt specific 3D tasks, as in Eq.~\ref{eq:pointadapter}.
\begin{equation}
  \textbf{h}^{PA} = f_{PA}(\textbf{h}^{P})
  \label{eq:pointadapter}
\end{equation}

Here we implement two versions of PointAdapter (PA), namely PTB-PA and FFN-PA, 
to handle the point cloud understanding tasks of different complexity. 

\noindent\textbf{i. PTB-PA}
is implemented as a Point Transformer~\cite{yu22pointbert} block (PTB), which stacks 
a multi-head self-attention (MSA) and a 2-layer MLP, added with the corresponding residual. 
For clarity, we denote this PointAdapter variant by Eq.~\ref{eq:ptb-pa}.
\begin{equation}
  \textbf{h}^{PA} = f_{PA}(\textbf{h}^{P}) = \textrm{PTB}(\textbf{h}^{P})
  \label{eq:ptb-pa}
\end{equation}

\noindent\textbf{ii. FFN-PA}
is designed as a Feed-Forward network (FFN) consisting of two Linear layers with GELU activation. 
It is equivalent to a residual MLP (ResMLP) submodule in the Point Transformer block. 
For simplicity, we use Eq.~\ref{eq:ffn-pa} to describe it. 
\begin{equation}
  \textbf{h}^{PA} = f_{PA}(\textbf{h}^{P}) = \textrm{ResMLP}(\textbf{h}^{P})
  \label{eq:ffn-pa}
\end{equation}

Now, we can predict a class distribution for point cloud $P$ by matching $\textbf{h}^{PA}$ with the 
text features $\textbf{h}^{T}_j$, $j\in 1,\dots, S$. The optimization procedure is introduced below. 

\subsubsection{Optimization}
The predicted class distribution for point cloud $P$ can be defined by 
Eq.~\ref{eq:text_pc_predcition_1} and Eq.~\ref{eq:text_pc_predcition_2}. 
\begin{equation}
  \hat{\textbf{y}} = [\hat{y}_1, \hat{y}_2, \dots, \hat{y}_S] \in \mathbb{R}^{S} \\
  \label{eq:text_pc_predcition_1}
\end{equation}
\begin{equation}\abovedisplayskip=0pt
  \hat{y}_j = \frac{\exp(s(\textbf{h}^{PA}, \textbf{h}_{j}^{T}))}{\sum_{k}^{S} \exp(s(\textbf{h}^{PA}, \textbf{h}_{k}^{T}))}, \quad j=1\dots S
  \label{eq:text_pc_predcition_2}
\end{equation}
$s(\cdot,\cdot)$ is the cosine similarity of the inputs and 
we exploit the cross entropy ($CE$) to compute the loss in Eq.~\ref{eq:ppt_loss}.
\begin{equation}
  \mathcal{L}_{CE} = \sum_i -\textbf{y}_i\log\hat{\textbf{y}}_i - (1-\textbf{y}_i)\log(1-\hat{\textbf{y}}_i)
  \label{eq:ppt_loss}
\end{equation}
$\textbf{y}_i$ is the real distribution of $i$th point cloud in the training set. 
The parameters in PromptLearner and PointAdapter are initialized with a Gaussian distribution $\sim (0, 0.02^2)$, 
then updated via gradient back propagation. 

\section{Experiments}
\label{sec:experiments}

In this section, we conduct systematic experiments to evaluate the proposed approach on multiple 3D tasks and justify vital design choices. 

\subsection{3D Point Cloud Recognition}
\label{subsec:3d_recognition}

Point cloud recognition is evaluated on two public datasets, ModelNet40~\cite{wu15modelnet} and ScanObjectNN~\cite{uy19sonn}. 
Note ScanObjectNN has three common splits: OBJ\_ONLY (OBJ), OBJ\_BG (BG), and PB\_T50\_RS (PB). 
The overall accuracy (OA) and number of learnable parameters (\#Params) are metrics of interest and the results are 
presented in Tab.~\ref{tab:cls_mn40_so_variants}. 

Our model has two variants: PPT-FFN and PPT-PTB. The former represents the model with FFN PointAdapter and the latter is the model 
with PTB PointAdapter. 
We compare them with representative methods that use supervised or unsupervised + full fine-tuning strategies. 

The results show PPT-FFN is competitive compared to previous strong baselines.  
Meanwhile, PPT-PTB not only attains new state-of-the-art performances on both datasets but also 
demonstrate excellent parameter efficiency. 
For instance, PPT-PTB achieves the same recognition accuracy as PointMLP on ModelNet40 while reducing the learnable parameters by 85.7\%.
On the three splits of ScanObjectNN, our PPT-PTB gets 93.1\%, 95,4\% and 89.1\% accuracy, 
outperforming previous state-of-the-art ACT by 1.2\%, 2.1\% and 0.9\% and saving 20.3M learnable parameters. 
Compared to ULIP, 
PPT-PTB improves ULIP by 2.7\% accuracy on ScanObjectNN (PB)
but uses 95\% fewer learnable parameters. 

\begin{table}[ht]
  \caption{Comparison of point cloud recognition on ModelNet40 and three splits of ScanObjectNN.
  OBJ: objects only. BG: objects with background. PB: objects with perturbations.}
  \label{tab:cls_mn40_so_variants}
  \centering
  \begin{tabular}{l r c c c c}
    \toprule
    \multirow{2}{*}{Method} & \#Params & MN40 & OBJ & BG & PB \\
    & (M) & (\%) & (\%) & (\%) & (\%) \\
    \midrule
    \multicolumn{6}{c}{\emph{supervised training}} \\
    PointNet~\cite{qi17pointnet} & 3.5 & 89.2 & 79.2 & 73.3 & 68.0 \\
    PointNet++~\cite{qi17pointnet2} & 1.5 & 90.7 & 84.3 & 82.3 & 77.9 \\
    PointCNN~\cite{li18pointcnn} & 0.6 & 92.2 & 85.5 & 86.1 & 78.5 \\
    SpiderCNN~\cite{xu18spidercnn} & -- & 92.4 & 79.5 & 77.1 & 73.7 \\
    DGCNN~\cite{wang19dgcnn} & 1.8 & 92.9 & 86.2 & 82.8 & 78.1 \\
    SimpleView~\cite{goyal21revisiting} & 0.8 & 93.0 & -- & -- & 80.5 \\
    MVTN~\cite{Hamdi21mvtn} & 3.5 & 93.5 & 92.3 & 92.6 & 82.8 \\
    PointMLP~\cite{ma22pointmlp} & 12.6 & 94.1 & -- & -- & 85.4 \\
    PointNeXt~\cite{qian22pointnext} & 1.4 & 93.2 & -- & -- & 87.7 \\
    \midrule
    \multicolumn{6}{c}{\emph{unsupervised pre-training + full fine-tuning}} \\
    OcCo~\cite{wang21occo} & 3.5 & 93.0 & 85.5 & 84.9 & 78.8 \\
    CrossPoint~\cite{afham22crosspoint} & 27.7 & 90.3 & -- & 81.7 & -- \\
    PointBERT~\cite{yu22pointbert} & 39.1 & 93.2 & 88.1 & 87.4 & 83.1 \\
    MaskPoint~\cite{liu22maskpoint} & 22.1 & 93.8 & 89.7 & 89.3 & 84.6 \\
    PointMAE~\cite{pang22pointmae} & 22.1 & 93.8 & 88.3 & 90.0 & 85.2 \\
    PointCMT~\cite{yan22pointcmt} & 12.6 & 93.5 & -- & -- & 86.4 \\
    PointM2AE~\cite{zhang22pointm2ae} & 12.9 & 93.4 & 88.8 & 91.2 & 86.4\\
    ACT~\cite{dong23act} & 22.1 & 93.7 & 91.9 & 93.3 & 88.2 \\  %
    ULIP(PointBERT)~\cite{xue23ulip} & 39.1 & 94.1 & -- & -- & 86.4 \\
    \midrule
    \multicolumn{6}{c}{\emph{parameter-efficient prompt tuning}} \\
    \textbf{PPT-FFN}(PointBERT) &  \textbf{1.2} & \textbf{93.0} & \textbf{92.6} & \textbf{93.3} & \textbf{86.5} \\   %
    \textbf{PPT-PTB}(PointBERT) &  \textbf{1.8} & \textbf{94.1} & \textbf{93.1} & \textbf{95.4} & \textbf{89.1} \\  %
    \bottomrule
  \end{tabular}
\end{table}

\subsection{Few-shot Learning}
\label{subsec:3d_fewshot}

\begin{figure}[t]
  \begin{center}
          \includegraphics[width=0.5\linewidth]{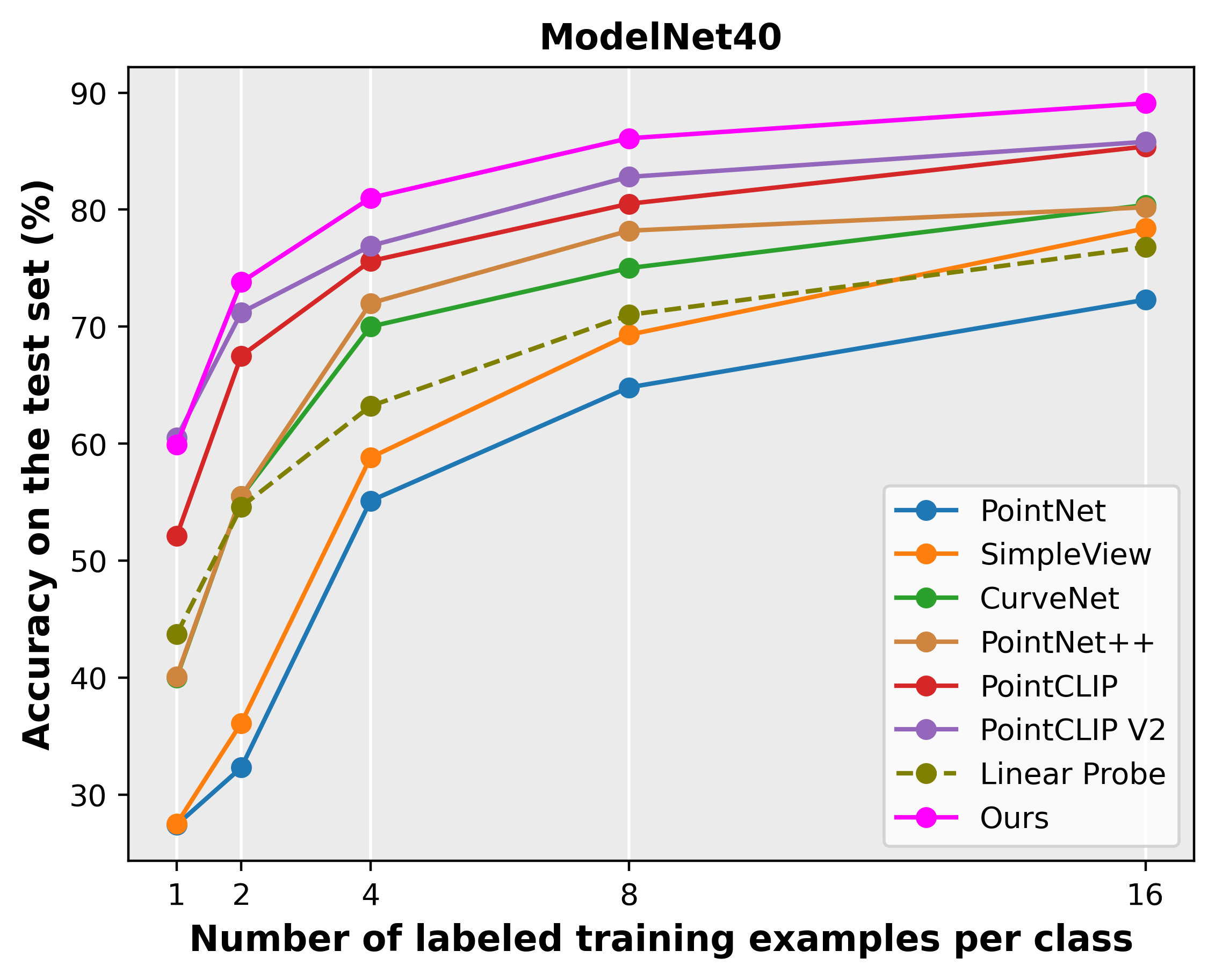}%
          \includegraphics[width=0.5\linewidth]{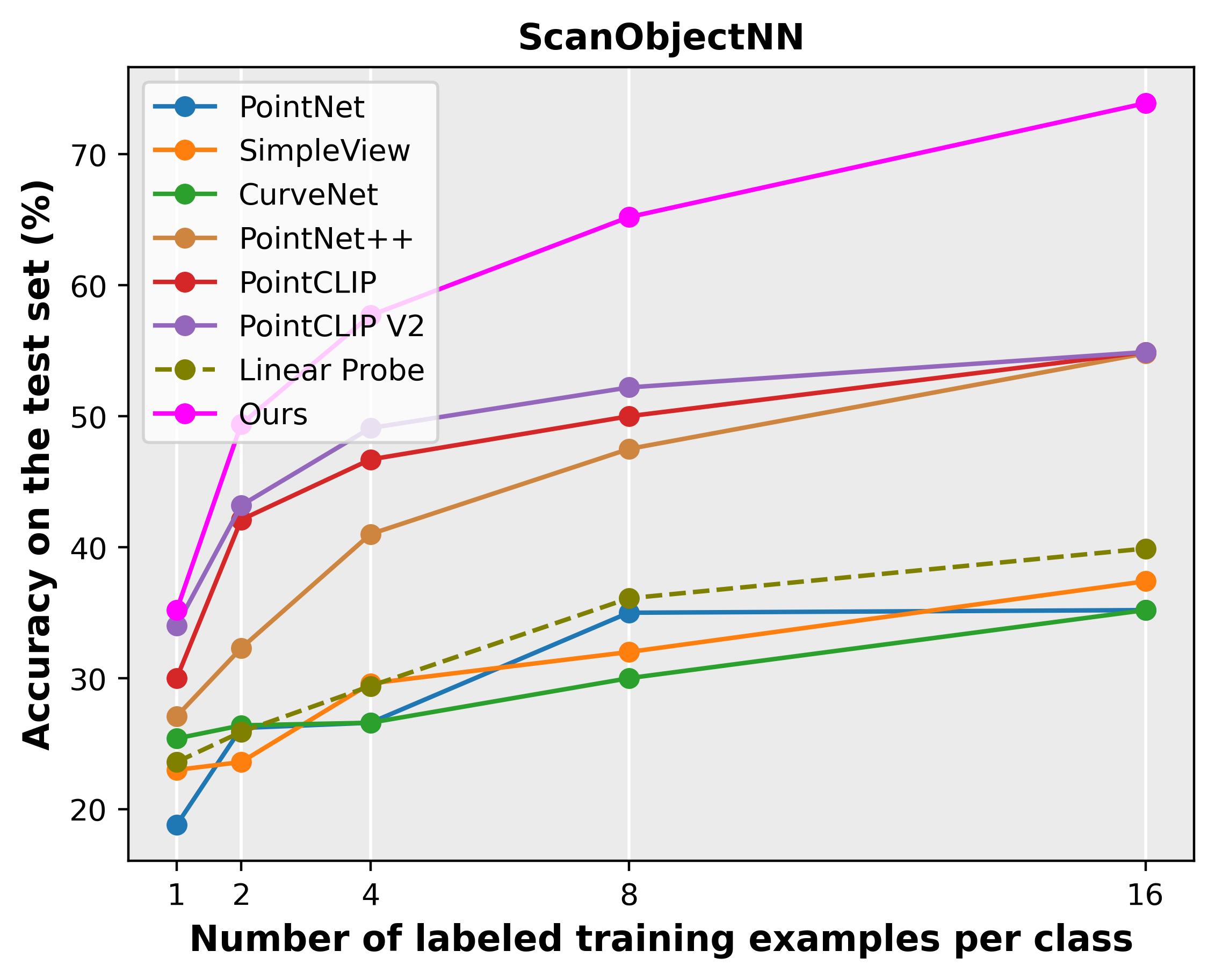}%
  \end{center}
  \caption{Comparison of few-shot classification of different methods on two datasets.}
  \label{fig:method_acc_params}
\end{figure}

We conduct few-shot point cloud classification on ModelNet40 and ScanObjectNN (PB). 
Following existing practices~\cite{zhang22pointclip,zhu23pointclip2}, 
1, 2, 4, 8, and 16 shots are randomly sampled from each category for training, but the evaluation takes place on the whole test set. 
We adopt PPT-FFN for experiments. 
The comparison with related methods is visualized in Fig.~\ref{fig:method_acc_params}. 
We re-implement PointCLIP V2~\cite{zhu23pointclip2} since there is no released code for the few-shot setting. 
Both PointCLIP~\cite{zhang22pointclip} and PointCLIP V2~\cite{zhu23pointclip2} use ResNet101~\cite{he2016resnet} as the backbone. 

The results demonstrate our model leads the runner up PointCLIP V2~\cite{zhu23pointclip2} by a clear margin on both datasets. 
The advantages are enlarged with increasing shots and difficulty of the downstream dataset.
Surprisingly, on the hardest split of ScanObjectNN (PB), PPT-FFN reaches 73.9\% accuracy using 16 shots, surpassing 
the runner up by 19\% absolute points. 
The experiments validate our parameter-efficient prompt tuning strategy makes ULIP a better 3D learner 
under a low-data regime. 

\subsection{3D Shape Part Segmentation}
\label{subsec:3d_partseg}

We conduct 3D shape part segmentation on the ShapeNetPart~\cite{shapenetpart} dataset. The major metrics for evaluation include 
overall accuracy (OA), mean class-wise intersection over union (mIoU$_C$), mean instance-wise interaction over union (mIoU$_I$) 
and number of learnable parameters (\#Params). 

For this task, we append a part segmentation head on the 3D encoder as in \cite{yu22pointbert,pang22pointmae,zhang22pointm2ae,
dong23act,sun23vipformer}. The results are displayed in Tab.~\ref{tab:partseg_shapenetpart}. 
Similarly, the proposed model outperforms the supervised and unsupervised counterparts, obtaining 86.4\% mIoU$_C$ and 88.1\% mIoU$_I$. 
Note prior competitive I2P-MAE~\cite{zhang23i2pmae} is also a multi-modal model, which 
converts point clouds into images to absorb off-the-shelf 2D knowledge~\cite{dosovitskiy21vit,radford21clip}. 
Instead, this work chooses to adapt the multi-modal ULIP by parameter-efficient prompt tuning, outreaching I2P-MAE 
by 1.2\% mIoU$_C$ and 1.3\% mIoU$_I$ with only 29\% parameters of it.

\begin{table}[ht]
  \caption{Comparison of 3D object part segmentation on ShapeNetPart.}
  \label{tab:partseg_shapenetpart}
  \centering
  \begin{tabular}{l c c c r }
    \toprule
    \multirow{2}{*}{Method} & OA & mIoU$_{C}$ & mIoU$_{I}$ & \#Params \\
    & (\%) & (\%) & (\%) & (M) \\
    \midrule
    \multicolumn{5}{c}{\emph{supervised training}} \\
    PointNet~\cite{qi17pointnet} & -- & 80.4 & 83.7 & 8.5 \\
    PointNet++~\cite{qi17pointnet2} & -- & 81.9 & 85.1 & 6.5 \\
    DGCNN~\cite{wang19dgcnn} & -- & 82.3 & 85.2 & 5.6 \\
    \midrule
    \multicolumn{5}{c}{\emph{unsupervised pre-training + full fine-tuning}} \\
    Transformer & -- & 83.4 & 85.1 & -- \\ %
    CrossPoint~\cite{afham22crosspoint} & 93.8 & 84.3 & -- & 27.5 \\
    PointBERT~\cite{yu22pointbert} & -- & 84.1 & 85.6 & 44.1 \\
    MaskPoint~\cite{liu22maskpoint} & -- & 84.4 & 86.0 & 27.1 \\
    PointMAE~\cite{pang22pointmae} & 94.8 & -- & 86.1 & 27.1 \\ %
    PointM2AE~\cite{zhang22pointm2ae} & 94.9 & 84.9 & 86.5 & 25.5 \\  %
    ViPFormer~\cite{sun23vipformer} & 94.8 & 84.7 & -- & 26.8 \\
    ACT~\cite{dong23act} & -- & 84.7 & 86.1 & 27.1 \\
    I2P-MAE~\cite{zhang23i2pmae} & -- & 85.2 & 86.8 & 17.9 \\ %
    \midrule
    \multicolumn{5}{c}{\emph{parameter-efficient prompt tuning}} \\
    \textbf{PPT}(PointBERT) & \textbf{95.0} & \textbf{86.4} & \textbf{88.1} & \textbf{5.2} \\ %
    \bottomrule
  \end{tabular}
\end{table}

\subsection{Data Efficiency}
\begin{figure}[t]
  \centering
  \begin{subfigure}{0.5\linewidth}
    \includegraphics[width=\linewidth]{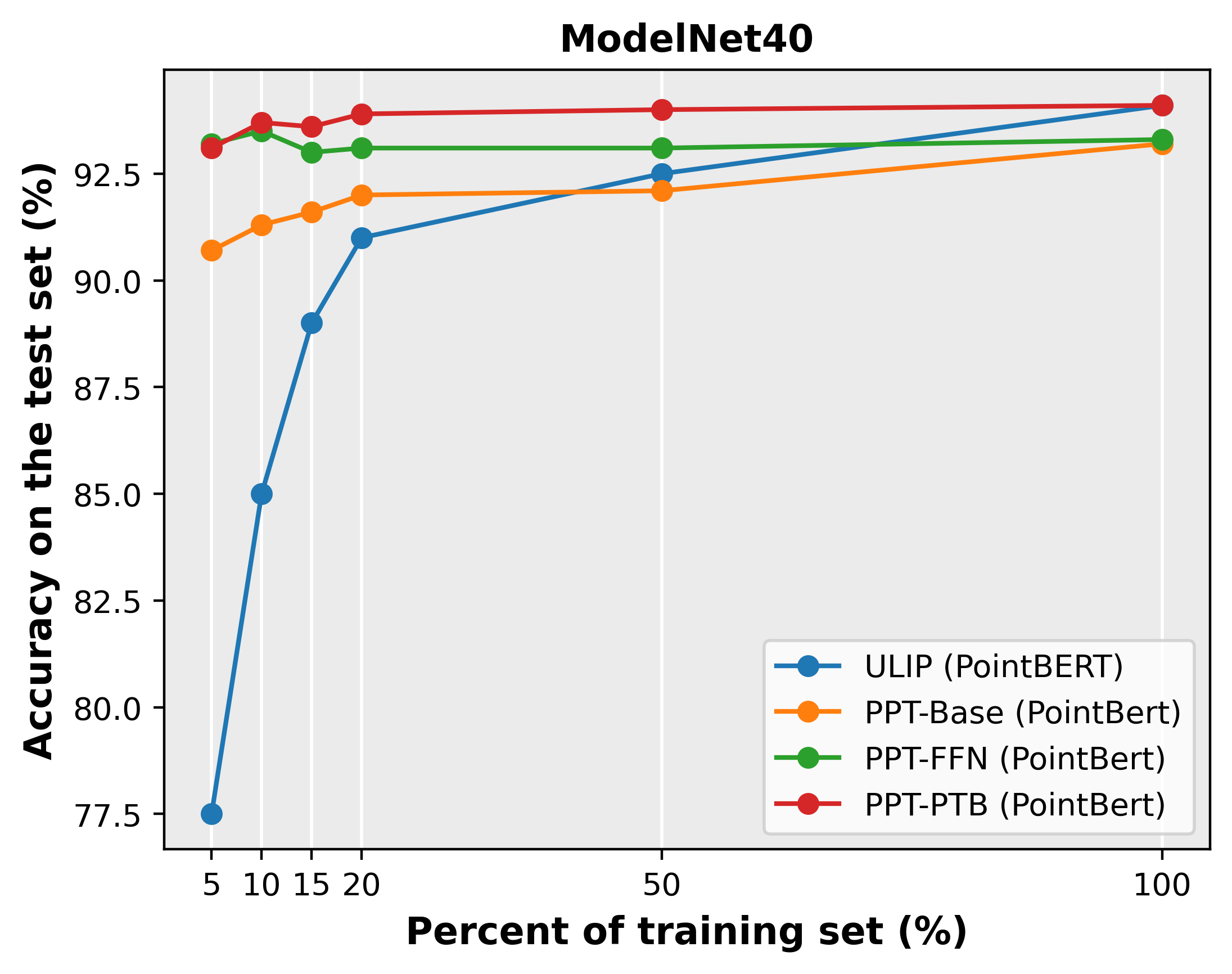}
    \caption{data efficiency}
    \label{fig:data_effi_mn40}
  \end{subfigure}%
  \begin{subfigure}{0.49\linewidth}
     \includegraphics[width=\linewidth]{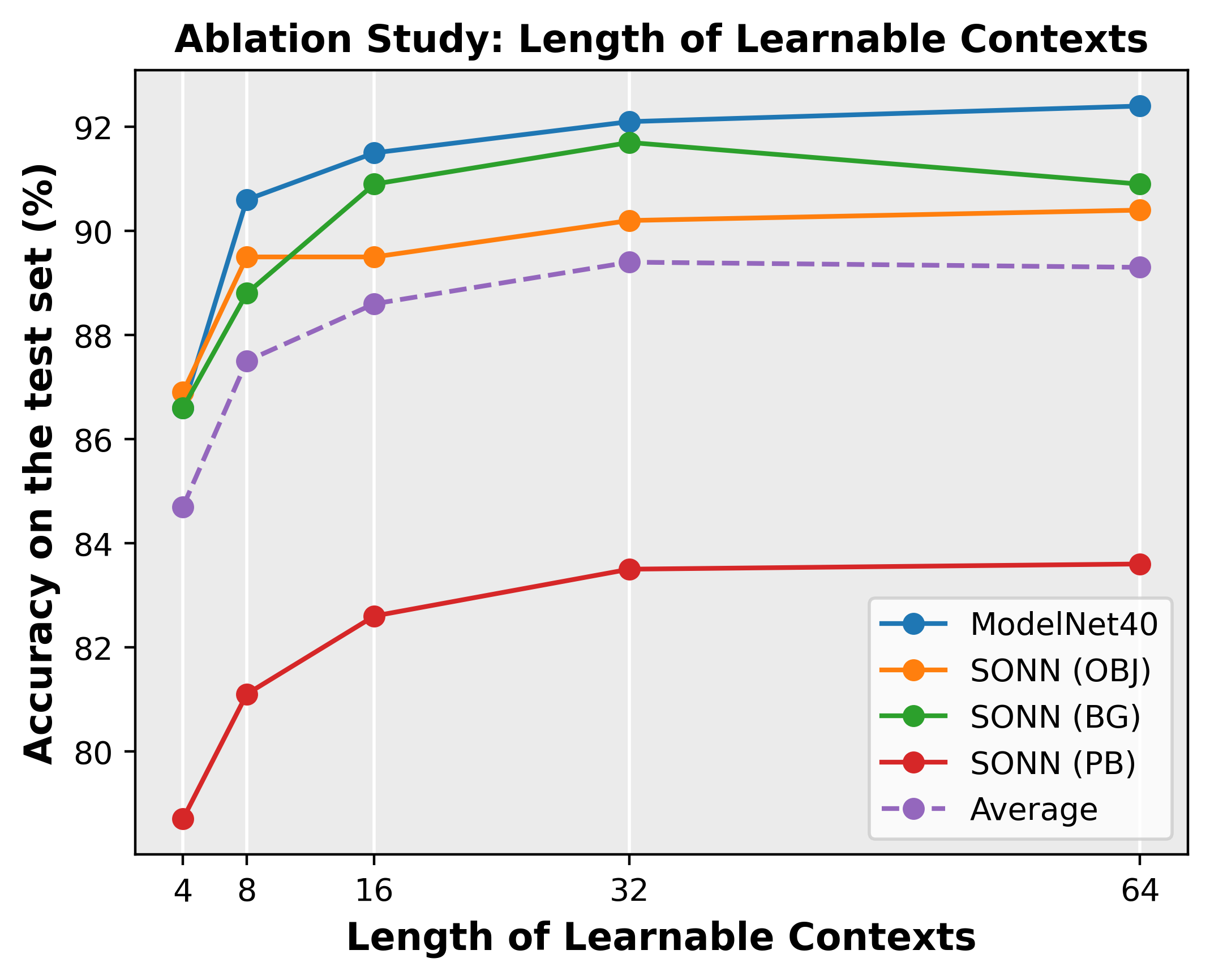}
     \caption{context length}
     \label{fig:ablate_context_len}
  \end{subfigure}
  \caption{In figure (a), the data efficiency between ULIP and PPT is compared. In figure (b), we ablate the context length
  on 4 datasets and the average is displayed in the dashed line.}
  \label{fig:data_effi_and_context_len}
\end{figure}

Adapting a large model to downstream tasks could potentially decrease the demand for labeled data. 
We investigate the data efficiency of the devised prompt tuning strategy and compare it with the full fine-tuning paradigm adopted by ULIP. 
The experiment is conducted on ModelNet40, using different portions (5\%, 10\%, 15\%, 20\%, etc.) of data for training and 
evaluating on the whole test set. Fig. \ref{fig:data_effi_mn40} exhibits the results. 
Here PPT-Base indicates our model only introduces the PromptLearner module, without PointAdapter. 
We observe under low-data regime, especially when using less than 20\% of training data, 
our three PPT variants lead ULIP (PointBERT) by significant margins. 
Even training with 5\% data and less than 1.8M learnable parameters, PPT-Base, PPT-FFN, and PPT-PTB reach 90.7\%, 93.2\% and 93.1\% test accuracy, 
respectively, versus 39.1M parameters and 77.5\% accuracy of ULIP. 
The results indicate the developed parameter-efficient prompt tuning strategy is also data-efficient. 

\subsection{Ablation Studies} 
\label{subsec:ablation}

We conduct a series of controlled experiments to examine the design choices of the proposed approach. 

\begin{table}[ht]
  \caption{\textbf{Ablation Study: Prompt Engineering vs. PromptLearner.}
  The prompt engineering combines 64 hand-crafted templates as in ULIP. 
  ``1k pts'' means a point cloud has 1024 points and ``8k pts'' are 8192 points.}
  \label{tab:ablate_soft_prompt_mn40_sonn}
  \centering
  \begin{tabular}{l | c c c | c}
    \toprule
    \multirow{3}{*}{3D Encoder} & ULIP & ULIP & \textbf{PPT-Base} & \multirow{3}{*}{$\Delta$} \\
    & Manual & Manual & Learnable & \\
    & (1k pts) & (8k pts) &  (1k pts) & \\
    \midrule
    & \multicolumn{3}{c|}{ModelNet40} \\
    PointNet++(SSG)~\cite{qi17pointnet2} & 55.6 & 57.7 & \textbf{89.2} & 33.6 \\ %
    PointNet++(MSG)~\cite{qi17pointnet2} & 58.4 & 55.9 & \textbf{88.7} & 30.3 \\ %
    PointMLP~\cite{ma22pointmlp} & 56.1 & 60.0 & \textbf{88.6} & 32.5 \\  %
    PointBERT~\cite{yu22pointbert} & 71.2 & 73.3 & \textbf{92.2} & 21.0 \\ %
    \midrule
    & \multicolumn{3}{c|}{ScanObjectNN} \\
    PointNet++(SSG)~\cite{qi17pointnet2} & 30.3 & 29.3 & \textbf{65.2} & 34.9 \\ %
    PointNet++(MSG)~\cite{qi17pointnet2} & 29.1 & 28.4 & \textbf{65.7} & 36.6 \\ %
    PointMLP~\cite{ma22pointmlp} & 30.3 & 30.1 & \textbf{63.3} & 33.0 \\  %
    PointBERT~\cite{yu22pointbert} & 33.2 & 37.2 & \textbf{83.6} & 50.4 \\   %
    \bottomrule
  \end{tabular}
\end{table}

\subsubsection{Prompt Engineering vs. PromptLearner}
In this work, we replace the manual prompts with learnable contexts. The manual prompts and learnable contexts are generated 
by prompt engineering and the PromptLearner module, respectively. 
The benefits of PromptLearner are verified in Tab.~\ref{tab:ablate_soft_prompt_mn40_sonn}. This table compares the recognition 
accuracy (in \%) under manual and learnable settings. The 2nd and 3rd columns record the results of zero-shot ULIP and the 4th column is ours. 
The last column is the improvement over ULIP (Manual, 1k pts). 
Note that the PPT-Base model is used and the performances are substantially boosted by deploying the PromptLearner module on the pre-trained ULIP. 
In most cases, the improvements are more than 30.0\% absolute points, up to 50.4\%. 
Also, the improvements can be generalized to different point cloud encoders (PointNet++, PointMLP, PointBERT) and 
datasets (ModelNet40 and ScanObjectNN). 

\subsubsection{Performance Gains brought by PointAdapter}
This experiment examines the performance gains brought by PointAdapter. 
The model for comparison is PPT-Base, which means there is no PointAdapter. 
Both PPT-FFN and PPT-PTB arrange the PointAdapter module. 
The recognition results in Tab.~\ref{tab:ablate_pointadapter} suggest 
the PPT variants with PointAdapter clearly improve PPT-Base (see $\Delta$). 

\begin{table}[ht]
  \caption{\textbf{Ablation Study: Gains brought by PointAdapter.} MN: ModelNet40. SO: ScanObjectNN.}
  \label{tab:ablate_pointadapter}
  \centering
  \begin{tabular}{l | c c | c c}
    \toprule
    Model & MN (\%) & $\Delta$ & SO (PB) (\%) & $\Delta$ \\
    \midrule
    \textbf{PPT-Base}(PointBERT) & 92.2 & -- & 83.6 & -- \\
    \midrule
    \textbf{PPT-FFN}(PointBERT) & \textbf{93.0} & 0.8 & \textbf{86.5} & 2.9 \\
    \textbf{PPT-PTB}(PointBERT) & \textbf{94.1} & 1.9 & \textbf{89.1} & 5.5 \\
    \bottomrule
  \end{tabular}
\end{table}

\subsubsection{The Length of Learnable Contexts}
One variable that should be decided is the length $M$ of the learnable contexts. 
Intuitively, longer contexts contain more parameters thus may provide the model with more informative descriptions for downstream tasks. 
We explore this problem by varying the length and comparing the recognition accuracy. 
The results are averaged over 4 datasets, referring to the dashed line in Fig.~\ref{fig:ablate_context_len}. 
The overall trend is the longer the context, the better the performance. But it is not always positive 
to increase length, i.e., PPT-Base of $M=64$ lags behind that of $M=32$ in average. Thus we adopt $M=32$ by default. 

\subsubsection{Template-based vs. Random Initialization}
Here we investigate different ways to initialize the learnable contexts. There are two modes: template-based and random. 
The first mode initializes the learnable contexts with the embeddings of a manual template, i.e., ``a point cloud model of a'', 
while the second one initializes them with random vectors. We compare the 3D classification accuracy (in \%) and 
the results are averaged on 4 datasets, including ModelNet40 and three splits of ScanObjectNN, 
shown in Tab.~\ref{tab:ablate_context_init}. 
In fact, there is no big difference between the two initializations. We adopt the random mode and middle class position by default. 

\begin{table}[ht]
  \caption{\textbf{Ablation Study: Template-based vs. Random Initialization} for the learnable contexts. 
  The 3D encoder in PPT-Base is PointBERT. Here front/middle/end indicates the inserted position of 
  a class name.}
  \label{tab:ablate_context_init}
  \centering
  \begin{tabular}{l | c c c | c c c }
    \toprule
    \multirow{3}{*}{Model} & \multicolumn{3}{c|}{Template-based} & \multicolumn{3}{c}{Random} \\
    & \multicolumn{3}{c|}{``a point cloud model of a''} & \multicolumn{3}{c}{$[\textbf{v}_1][\textbf{v}_2][\textbf{v}_3][\textbf{v}_4][\textbf{v}_5][\textbf{v}_6]$}\\\cline{2-4}\cline{5-7}
    & front & middle & end & front & middle & end \\
    \midrule
    \textbf{PPT-Base} & 87.11 & 87.13 & 83.53 & 87.11 & 87.13 & 83.53 \\
    \bottomrule
  \end{tabular}
\end{table}

\subsection{Visualization} 
\label{subsec:vis}

\noindent\textbf{Learned Contexts.}
The learned prompts are relatively hard to understand since they 
probably cannot be mapped to the words in a vocabulary. We try to interpret the learned prompts by finding their nearest words in a 
vocabulary based on Euclidean distance. The vocabulary uses BPE encoding~\cite{sennrich16nmt} as in CLIP~\cite{radford21clip}. 
For clarity, the length of learnable contexts is set to 6. After optimization on downstream datasets, the closest word for each learned vector is  
shown in Tab.~\ref{tab:vis_interprete_learned_prompts}. We observe the returned terms are not closely related to the 3D topics, 
and cannot form a meaningful sentence for human beings. Similar observations also occur in another work~\cite{zhou22coop}. 
It may be inappropriate to explain the learned contexts with nearest words. The problem is interesting and deserves further investigation. 

\begin{table}[ht]
  \caption{The nearest word for each of the 6 learned context vectors. 
  The number below the word is the distance between the learned context vector and its nearest word embedding in the vocabulary.}
  \label{tab:vis_interprete_learned_prompts}
  \centering
  \begin{tabular}{l | c c c c c c }
    \toprule
    No. & 1 & 2 & 3 & 4 & 5 & 6 \\
    \midrule
    \multirow{2}{*}{MN~\cite{wu15modelnet}} & bharti & etv & ihear & awaz & luhan & cnn \\
    & 1.729 & 1.676 & 1.589 & 1.619 & 1.605 & 1.694 \\
    \midrule
    \multirow{2}{*}{SO~\cite{uy19sonn}} & chatur & appear & letit & matil & smack & antino \\
    & 1.484 & 1.433 & 1.382 & 1.444 & 1.440 & 1.420 \\
    \bottomrule
  \end{tabular}
\end{table}

\begin{figure}[ht]
  \centering
  \includegraphics[width=\linewidth]{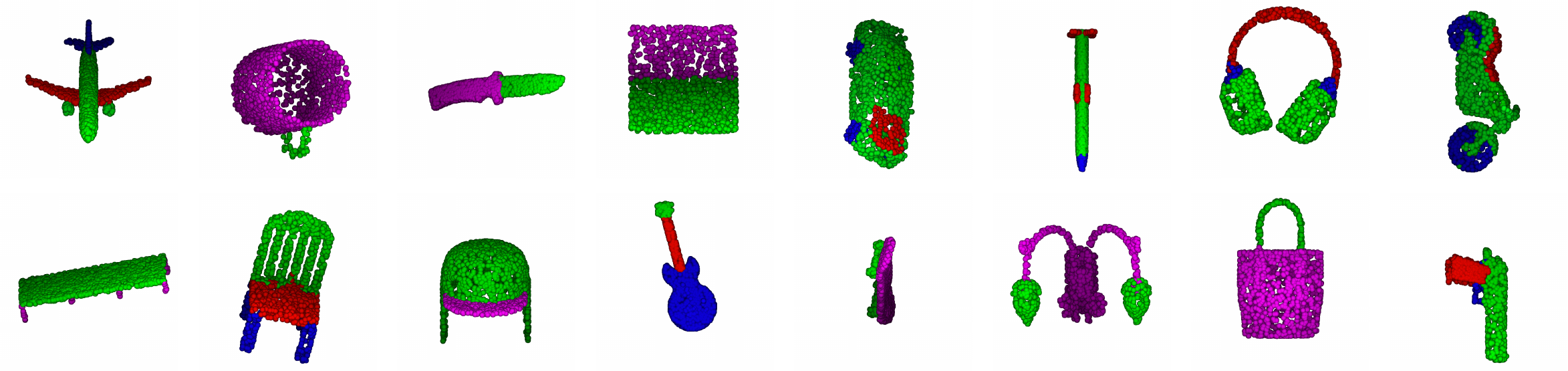}
  \caption{Part segmentation visualization for PPT predictions.} 
  \label{fig:vis_partseg}
\end{figure}

\noindent\textbf{3D Part Segmentation.} 
We visualize the part segmentation predictions of PPT on ShapeNetPart~\cite{shapenetpart}, which contains 16 classes. 
A single point cloud is randomly selected from each class for test. The different part predictions are mapped to different
colors for each 3D shape. The results in Fig.~\ref{fig:vis_partseg} indicate our model can segment object parts in
various categories accurately. 

\section{Conclusion}
\label{sec:conclusion}
In this paper, we develop a parameter- and data-efficient prompt tuning strategy to adapt a large multi-modal model for 3D point cloud understanding. 
A PromptLearner module is proposed to to elicit the rich knowledge encoded in the large model instead of depending on 
hand-crafted prompts. 
Based on that, we arrange a PointAdapter module near downstream tasks to further strengthen prompt learning. 
During optimization, the pre-trained 3D encoder is frozen and only parameters in PromptLearner and PointAdapter are updated. 
Experiments on various 3D tasks demonstrate the superior parameter and data efficiency of 
the proposed model, accompanied by record-breaking performances. 

\addtolength{\textheight}{-2cm}   %

\bibliographystyle{IEEEtran}
\bibliography{IEEEabrv}

\end{document}